\documentclass[sigplan,screen]{acmart}
\newcommand{\modelname}{MiTREE }
\AtBeginDocument{%
  }

\setcopyright{none}
\renewcommand\footnotetextcopyrightpermission[1]{}
\setcopyright{none}
\copyrightyear{2024}
\acmYear{2024}
\acmDOI{}
\acmConference[]{}{}{}
\acmISBN{}

\usepackage{pifont} 

\begin{document}

\title[MiTREE]{MiTREE: Multi-input Transformer Ecoregion Encoder for Species Distribution Modelling}


\author{Theresa Chen}
\affiliation{%
  \institution{University of Minnesota}
  \city{Minneapolis}
  \country{Minnesota}}
\email{chen7924@umn.edu}

\author{Yao-Yi Chiang}
\affiliation{%
  \institution{University of Minnesota}
  \city{Minneapolis}
  \country{Minnesota}}
\email{yaoyi@umn.edu}


\begin{abstract}
  Climate change poses an extreme threat to biodiversity, making it imperative to efficiently model the geographical range of different species. The availability of large-scale remote sensing images and environmental data has facilitated the use of machine learning in Species Distribution Models (SDMs), which aim to predict the presence of a species at any given location. Traditional SDMs, reliant on expert observation, are labor-intensive, but advancements in remote sensing and citizen science data have facilitated machine learning approaches to SDM development. However, these models often struggle with leveraging spatial relationships between different inputs- for instance, learning how climate data should inform the data present in satellite imagery- without upsampling or distorting the original inputs. Additionally, location information and ecological characteristics at a location play a crucial role in predicting species distribution models, but these aspects have not yet been incorporated into state-of-the-art approaches. In this work, we introduce MiTREE: a multi-input Vision-Transformer-based model with an ecoregion encoder. \modelname computes spatial cross-modal relationships without upsampling as well as integrates location and ecological context. We evaluate our model on the SatBird Summer and Winter datasets, the goal of which is to predict bird species encounter rates, and we find that our approach improves upon state-of-the-art baselines.
\end{abstract}



\maketitle

\section{Introduction}
The escalating effects of climate change pose serious threats to biodiversity through habitat loss and rapid changes within ecosystems. Biodiversity is essential for maintaining ecosystem services that benefit human well-being, such as pollination and nutrient cycling \cite{haines2010links}. Planning for biodiversity conservation requires us to efficiently model areas that species' are currently occupying. However, traditional methods to create species distribution models (SDMs) primarily rely on expert observation and knowledge, making the process very time and labor-intensive. Recently, the increasing availability of remote sensing images and citizen science data has led to the development of large-scale datasets that can facilitate machine learning methods for creating SDMs \cite{teng2024satbird, lorieul2022overview}. 

Prior approaches have begun to harness the vast amount of information in large-scale species distribution datasets and have developed deep learning models that are customized for the task of species distribution modeling. For instance, several studies have utilized geographic locations as an indicator to predict species occurrence vectors \cite{cole2023spatial, lange2024active}. However, relying solely on geographic location may not provide accurate species distributions across diverse species, as environmental factors, such as climate or geography can influence spatial patterns \cite{hirzel2008habitat}. To achieve robust predictions and capture the full spectrum of influences on species distribution, models must be able to effectively integrate multi-modal geographic data.

Currently, state-of-the-art (SOTA) methods that integrate multiple sources are computer vision models that have been adapted from their original source domains for species distribution modeling. For instance, studies have adapted ResNet to integrate red-green-blue (RGB) satellite imagery, the near-infrared (NIR) channel, and environmental data \cite{teng2024satbird, lorieul2022overview} for the purpose of representation learning for a geographic area. These approaches require all modalities to be upsampled and aligned to a common resolution before they are concatenated and passed into the model. Upsampling prior to being passed through the neural network can affect the robustness of the representation as it sacrifices the fidelity of the original input and increases noise. Some approaches avoid upsampling by passing each input through an individual convolutional network and concatenating the learned representations for each input afterward \cite{lorieul2022overview}. However, these models do not explicitly account for spatial cross-correlations between different data sources. For example, given a patch of trees in satellite imagery, the information in the corresponding patch in the climate data may help explain what kind of forest is present. Models that have been trained to quantify cross-modal relations between data from different sources have the potential to outperform baseline models on tasks that require robust representation for geographic areas.

There exist several SOTA vision models that can perform multimodal learning with attention without upsampling \cite{bachmann2022multimae, Girdhar_2023_CVPR}. These models assume that images from different modalities have similar, if not the same, image dimensions (number of pixels on each side). However, when dealing with geographic data, one input's image dimensions can be 20 to 30 times larger than another's. Our aim is to utilize the base structures of SOTA multimodal vision models but adapt the architecture to account for inputs of vastly different resolutions.

One challenge unique to geographic data is representing the geographic location of the data. Location is imperative in identifying underlying patterns in spatial data. Species data are particularly sensitive to geographic position, as movement limitations, such as maximum flight or foraging distances, often restrict species' ranges. In the past, machine learning models have used location information to augment representations for classification tasks with a geographic component \cite{mai2023sphere2vec} or clustered location information as a pre-text task during pre-training \cite{ayush2021geography}. Using raw latitude and longitude values can present several issues, such as bias due to not accounting for the earth's curvature and being too noisy \cite{mai2022review}. Having broader, region-level labels can help reduce the complexity of the location encoding task and simplify the model's process for learning spatial relationships, as it now deals with a smaller number of labels rather than a vast range of raw coordinates. 

In our proposed model, we incorporate a map defining ecological regions in the contiguous U.S., compiled by domain experts \cite{omernik2014ecoregions}, in order to make the model location-aware. By using ecoregions as location indicators, we address a common limitation of purely geographic clustering: locations that are close spatially but differ ecologically may still be grouped together if location alone is used. Ecoregions, however, are defined by shared environmental characteristics such as climate, vegetation, and topography \cite{omernik2014ecoregions}. This approach ensures that regions are grouped not only by proximity but also by ecological similarity, allowing the model to better capture the environmental context necessary for species distribution predictions.

For the task of species distribution modeling, we propose MiTREE, a multi-input ecoregion-aware vision Transformer (ViT) model that computes the attention between inputs of multiple resolutions without upsampling. We test our model on the species distribution modeling dataset SatBird, which utilizes imagery and environmental data of highly varied ground sampling distances (GSDs) (from 10 meters per pixel to 1,000 meters per pixel) to predict bird species distributions in the United States.

Our contributions are as follows:
\begin{itemize}
    \item First, we present \modelname, which builds upon the architecture of a multi-input ViT to jointly train satellite imagery and environmental data to predict species distributions across the United States. \modelname introduces two key modifications to the original multi-input ViT. The first is modified patch projection layers tailored to the data present and number of pixels in each input, which circumvents the needs for upsampling and enhances the representation quality of the inputs. The second is an ecoregion encoder that embeds geographic and ecological context directly into the input embeddings.
    \item We show that our model outperforms the existing SOTA baselines over the SatBird Summer and Winter datasets. 
    \item Through comprehensive ablation studies, we highlight the importance of each modification made to the original multi-input ViT structure in \modelname.
\end{itemize}

\section{Related Work}
\subsection{Species Distribution Models}
Species distribution modeling is the process of predicting where species are likely to occur by analyzing species occurrence data and relevant predictors, such as environmental variables \cite{elith2009species}. Traditional methods for SDMs encompass a range of statistical and machine learning methods. Commonly used models include logistic regression \cite{ferner2002extended}, spatial statistics \cite{doser2023joint, rangel2006towards}, regression trees \cite{elith2008working, prasad2006newer}, and maximum entropy models \cite{phillips2006maximum}. 

Several studies have applied deep learning to species distribution modeling, with a range of input types across different approaches. For example, Botella et al. implement a convolutional network to process environmental data \cite{botella2018deep}. The MOSAIKS method offers a low-computation machine learning method by using a single convolutional layer on satellite imagery and combining the resulting representations with tabular environmental data \cite{rolf2021generalizable}. Other approaches rely primarily on geographic location as the input \cite{cole2023spatial}. Some machine learning models distill satellite imagery into tabular features such as greenness and brightness \cite{seo2021stateconet}, while others jointly train ground-level bird images with satellite imagery \cite{sastry2024birdsat} through contrastive learning. However, none of these models jointly train environmental data and satellite imagery together and compute the spatial cross-correlations between the modalities.

\subsection{Multimodal Vision Models}
Our work builds upon existing multimodal vision models with a Transformer backbone. Although there exist several models for multimodal data combinations \cite{zhang2023cmx, Girdhar_2023_CVPR, zhang2023delivering}, this work adapts the architecture of MultiMAE, a self-supervised masked autoencoder ViT model. As MultiMAE was developed for self-supervised learning, it's architecture is not adapted for any particular downstream task and the structure of the multimodal ViT allows us to combine all input sources without any upsampling operations. MultiMAE jointly trains three different types of image inputs (RGB images, segmentation masks, and depth maps) by tokenizing each through a patch projection layer and then computing attention between each token in a Transformer. While the base architecture of MultiMAE is very useful, we have to modify the patch projection layers due to the varying resolutions of our input data and the addition of a location encoder. Further information on the adapted MultiMAE structure is detailed in section 3.2. 

\subsection{Geographic Foundation Models}
Geographic foundation models are neural networks pretrained on large datasets to learn generalized representations for geographic areas. These representations can then be finetuned or used out of the box for a range of downstream tasks. While not explicitly trained for the species distribution modeling task, these models can be finetuned on a species dataset and converted into SDMs.

Foundation models often address issues unique to geographic data. For instance, ScaleMAE and Cross ScaleMAE incorporate the GSD of the satellite image into the positional encoding of the Transformer \cite{reed2023scale, tang2024cross} to address the problem of images having different GSDs. Other models also utilize the Transformer structure but focus on handling multi-spectral imagery and dealing with temporal discrepancies between images \cite{cong2022satmae}. Several models use a ResNet \cite{he2016deep} structure and contrastive learning approach based on the geographic or temporal distance between satellite images \cite{ayush2021geography, manas2021seasonal}. While these models provide good generalized representations, they focus on learning representations for satellite imagery and do not explicitly incorporate environmental data, which is essential for SDMs.

\section{MiTREE} 
\begin{figure}[b]
  \centering
  \vspace{5mm}
  \includegraphics[width=\linewidth]{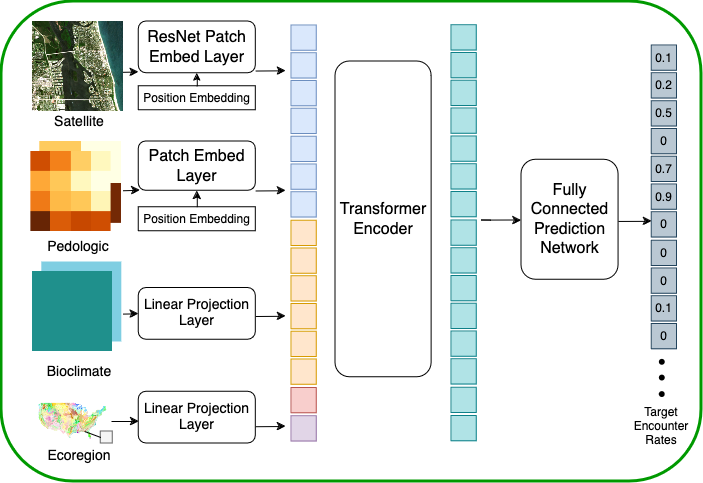}
  \caption{Overview of the \modelname architecture.}
\end{figure}
\subsection{Problem Setting}
    Our goal is to be able to predict bird sightings using remote sensing and environmental covariates as input data. Let $h$ be a geographic area of interest, which we will refer to as a hotspot, and $s_1, ..., s_n$ be our target species. At each $h$, our aim is to predict $y^h = (y^h_{s_1}, ...y^h_{s_n})$ where each $y^h_{s_1}$ represents the encounter rate for that hotspot. The encounter rate reflects the probability that a bird will be spotted if someone is in the hotspot and reflects the number of times a bird was sighted by citizen scientists at the target hotspot.

\subsection{\modelname Encoder Overview}
    The base of the encoder in our model is the ViT \cite{dosovitskiy2020image}. A vanilla Transformer encoder traditionally uses a single patch projection layer to split each $H, W$ input image into $n x m$ patches. These $n x m$ patches are then concatenated sequentially, and each patch is treated as a token input for the Transformer model. To preserve the knowledge of the spatial position before the Transformer encoder, a position embedding indicating the 2-D pixel position of the patch is added to the tokenized input. 
    
    To incorporate multiple data sources, we follow an architecture similar to MultiMAE \cite{bachmann2022multimae}, using a separate patch projection layer for each input type as well as separate positional embeddings for the tokens of each input. However, while MultiMAE uses a single layer convolutional network for each patch projection layer, \modelname implements different types of networks for each patch projection layer, depending on the data type and image dimensions of the original input. The specific architecture of each patch projection layer is detailed in Section 3.3. Additionally, because MultiMAE was not originally designed for geographic data, it does not natively support any way to integrate geographic location into the hotspot representation. \modelname includes a location encoder that tokenizes a unique location identifier called an ecoregion category, allowing the model to have information about the general location of the hotspot.

    After \modelname tokenizes all the inputs and the location, they are concatenated together and passed through a Transformer layer. In this layer, an encoder module comptues attention between all the tokens, allowing the model to make cross-connections between inputs from different sources. 

\subsection{Patch Embeddings}
Each hotspot in the SatBird dataset has an approximate geographic area of 640 meters squared ($m^2$) and has three raster data source types, as described below: 
\begin{itemize}
    \item \textbf{Satellite Imagery.} Sampled at 10 m/pixel with an image size of 64 x 64 pixels. Tokenized by a ResNet patch embedding layer, described below. 
    \item \textbf{Pedologic Data.} Sampled at 100 m/pixel with an image size of 4 x 4 pixels. Tokenized by a patch embedding layer that is comprised of a single convolutional layer.
    \item \textbf{Bioclimate Data.} Sampled at 1,000 m/pixel with an image size of 1 x 1 pixels. Tokenized by a single-layer linear projection network.
\end{itemize}
Having separate projection layers for each data source allows us to tokenize the inputs without having to upsample or alter the resolution of the original data. This approach guarantees that coarser inputs are not overrepresented, reducing noise and conserving memory within the model. 

For the satellite imagery input, the patch embedding layer is a ResNet18 \cite{he2016deep} rather than the typical single-layer convolutional network. There are several advantages to using ResNet for satellite imagery. Firstly, ResNet allows for deeper network training and condenses the image into a few patches with many channels. Satellite images are an inherently noise-heavy medium with many redundant pixels; for instance, in green areas like forests or grasslands, large portions of the imagery may consist of homogeneous regions with little variation. These images in particular may benefit from the greatly benefit from the ResNet progressively reducing the image dimension, as it helps reduce noise but still retain key features. Moreover, a more powerful patch embedding layer enables \modelname to establish a more semantically meaningful representation of the imagery. Finally, pretrained ResNet weights are widely available and can be used to create a 'warm start' effect where weights are already biased toward image data, accelerating model convergence \cite{ash2020warm}.

The pedological data and bioclimate data both have single-layer patch embeddings because their image dimensions are so small that a single convolutional or linear layer suffices. While they have a higher amount of channels (8 for the pedologic data and 19 for the bioclimate data), these are sufficiently captured by a shallower network due to the small number of pixels required to cover the entire hotspot area. For the pedologic data, because the image size is 4 x 4 pixels and the pixels still have spatial information that must be retained, we use a convolutional layer. For the bioclimate data, because there is only 1 pixel to represent the whole hotspot, there are no spatial correlations within the raster and it is sufficient to use a linear projection network. Additionally, there are no pretrained models for these two data sources, so the ResNet would not benefit from any 'warm start' effect.

\subsection{Ecoregion Location Encoder}
Ecoregions are areas characterized by similar abiotic and biotic conditions, including but not limited to geology, soil, climate, vegetation, and wildlife activity \cite{omernik2014ecoregions}. Ecoregions are location based; in other words, two regions far from each other with similar climatic conditions will not be classified as the same. The ecological map used in \modelname was created by domain experts working with the Environmental Protection Agency as part of a U.S.-wide mapping project \cite{mcmahon2001developing}. 

\begin{figure}[h]
  \centering
  \includegraphics[width=\linewidth]{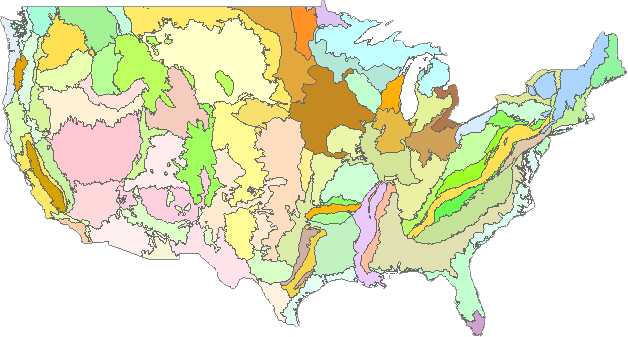}
  \caption{Map of Level III ecoregions in the United States. Each ecoregion is a unique spatial location with a categorical label and is represented by a different color on the map. }
  \label{fig:ecoregion_map}
  \Description{}
\end{figure}

There are several levels of ecoregion maps, with Ecoregion I having the largest areas at the coarsest level and Ecoregion IV having the smallest areas at the most fine-grained level. We use the labels from the Ecoregion III map (Figure \ref{fig:ecoregion_map}) because it strikes a balance between maintaining generalizability for birds with large ranges and capturing ecological differences between regions. 

In \modelname, the ecoregion label is passed through a single linear layer, and the resulting representation is treated as a token that is concatenated to the other tokens from the other input modalities before the tokens are passed through the Transformer.
\subsection{Position Embeddings}
For input types with a 2D structure (satellite and pedological), we apply a 2D sine-cosine position embedding after the data passes through the patch embedding layer \cite{he2020momentum}. 
The equations for calculating the positional embeddings are as follows:
\begin{align}
    & X_{(pos, 2i)} = sin(\frac{pos}{10000^{\frac{2i}{D}}}) .\label{eq:1}
\end{align}
\begin{align}
    & Y_{(pos, 2i + 1)} = cos( \frac{pos}{10000^{\frac{2i}{D}}}).\label{eq:2}
\end{align}
where \emph{pos} is the position of the patch, \emph{i} is the feature dimension, and \emph{D} is the dimension of the feature vector. 

\begin{table*}[h]
\centering
\caption{Results on the test split of SatBird USA Summer Dataset. Bold results are the best performing, and underlined results are the second best. An asterisk (*) denotes model results sourced from the SatBird dataset paper \cite{teng2024satbird}; all other results were generated using our computational resources. E2E indicates the model underwent end-to-end training. For MAE and MSE, a lower number is better but for top-10, top-30, and top-k, a higher number is better.}
\label{tab:summer}
\begin{tabular}{lccccccc}
\toprule
Model & Pretraining & MAE[1e-2] & MSE[1e-2] & Top-10 & Top-30 & Top-k \\
\midrule
GBRT* & - & 2.3 & .9 & 26.9\% & 38.6\% & 44.8\% \\
MOSAIKS* & - & 2.5 & .7 & 41.9\% & 56.7\% & 62.2\% \\
ResNet18 & ImageNet & \underline{2.134} & \underline{.642} & \underline{46.561\%} & \underline{65.744\%} & \underline{67.121\%} \\
SatMAE & fMoW-RGB & 3.101 & .939 & 28.550\% & 43.038\% & 46.113\% \\
SatMAE (img + env) & fMoW-RGB & 2.249 & .692 & 42.647\% & 62.290\% & 64.315\% \\
SATLAS & Sentinel-2 & 2.982 & .936 & 29.201\% & 45.623\% & 46.774\% \\
SATLAS (E2E) & Sentinel-2 & 2.948 & .914 & 29.708\% & 44.783\% & 47.728\% \\
\midrule
\modelname & ImageNet & \textbf{2.070} & \textbf{.630} & \textbf{47.380\%} & \textbf{66.609\%} & \textbf{67.821\%} \\
\bottomrule
\end{tabular}
\end{table*}

\begin{table*}[h]
\centering
\caption{Results on the test split of SatBird USA Winter Dataset. Bolded results are the best performing and underlined results are the second best. An asterisk (*) denotes model results sourced from the SatBird dataset paper \cite{teng2024satbird}; all other results were generated using our computational resources. E2E indicates the model underwent end-to-end training. For MAE and MSE, a lower number is better but for top-10, top-30, and top-k, a higher number is better.}
\label{tab:winter}
\begin{tabular}{lccccccc}
\toprule
Model & Pretraining & MAE[1e-2] & MSE[1e-2] & Top-10 & Top-30 & Top-k \\
\midrule
GBRT* & - & 2.5 & .7 & 27.6\% & 45.7\% & 51.4\% \\
MOSAIKS* & - & 1.9 & .5 & 47.8\% & 62.1\% & 66.4\% \\
ResNet18 & ImageNet & \underline{1.658} & \underline{.450} & \underline{51.416\%} & \underline{69.560\%} & \underline{70.955\%} \\
SatMAE* & fMoW-RGB & 2.4 & .7 & 28.6\% & 50.2\% & 52.6\% \\
SatMAE (img + env) & fMoW-RGB & 1.745 & .469 & 48.269\% & 67.099\% & 69.009\% \\
SATLAS* & Sentinel-2 & 2.3 & .7 & 31.6\% & 51.5\% & 54.1\% \\
SATLAS (E2E) & Sentinel-2 & 2.361 & .674 & 30.805\% & 51.080\% & 53.762\% \\
\midrule
\modelname & ImageNet & \textbf{1.621} & \textbf{.430} & \textbf{51.773\%} & \textbf{70.005\%} & \textbf{71.401\%}
 \\
\bottomrule
\end{tabular}
\end{table*}

\subsection{Prediction Layer}
Once a representation is obtained from the MultViT encoder, it is passed through a fully connected network to obtain a final prediction. The model is trained through Cross Entropy Loss, described below:
\begin{align}
     & \text{Loss (Cross Entropy)} = \frac{1}{N_h} \sum_{h} \text{L}_h \\
     & \text{L}_h = \sum_{s} \left[ -y_s^h \log(\hat{y}_s^h) - (1 - y_s^h) \log(1 - \hat{y}_s^h) \right]\label{eq:2}
\end{align}
Where $N_h$ is the number of hotspots, $h$ is the hotspot, $y$ is the predictions, $s$ is the species, and $\hat{y}$ is the ground truth.

\section{Experiments and Results}
\subsection{Experimental Settings}
The \modelname model is trained with a Transformer encoder with a dimension size of 512 and 8 heads. Each head has 12 attention layers. The batch size is 128, the learning rate is 0.0001, the optimizer is Adam, and the scheduler reduces the learning rate upon a plateau. The model was trained on a single 40GB A100-8. 

We train and evaluate our model on the SatBird Summer (Table 1) and SatBird Winter (Table 2) datasets using the task of species encounter rate for 670 bird species across the United States. For SatBird Summer there are 104,064 hotspots used during training and 18,529 hotspots used for testing. For the winter split, there are 46,564 training hotspots and 6797 testing hotspots. Summer and winter splits are separated because bird species distributions change significantly between seasons.

We perform an evaluation using five metrics: Mean Average Error (MAE), Mean Standard Error (MSE), top-10, top-30, and top-k (Table 1), following the evaluation method from the original SatBird paper \cite{teng2024satbird}. MAE and MSE measure the accuracy of the encounter rate prediction. The top-10 and top-30 metrics measure whether the species with the top 10 or 30 predicted encounter rates are also the top 10 or 30 observed species at the hotspot. The top-k metric measures, for the k-species at the hotspots with a non-zero encounter rate, whether the species with the top-k predicted encounter rates are the same.

The top-10, top-30, and top-k accuracy metrics are valuable because they will not be affected by the zero-inflation effects. Zero inflation is common in ecological data where species may be absent from many locations, resulting in numerous hotspots where the majority of the species ground truth encounter rates are zero. This naturally results in lower MSE values since the models can trend toward lower values and obtain low errors for many species at many hotspots. 

\subsection{Comparison to Baselines}
We compare our model to the baselines tested in the original SatBird dataset paper \cite{teng2024satbird} and introduce a few additional methods for comprehensive benchmarking. Our aim is to capture a wide range of models, from machine learning methods that might be used by ecologists or other domain experts to SOTA machine learning methods that could conceivably be adapted for the species modeling task. The baselines are described below:
\begin{itemize}
    \item \textbf{Gradient Boosted Regression Trees (GBRT).} GBRT models have been used in species distribution modeling among domain experts and are computationally efficient \cite{hao2020testing}. We report results from the original SatBird paper, which applies GBRT to bioclimatic and pedological data.
    \item \textbf{MOSAIKS.} Originally designed for computational efficiency and generalizability \cite{rolf2021generalizable}, MOSAIKS uses fixed random filters on input images, combining these outputs with environmental variables in a regressor to predict encounter rates at each hotspot.
    \item \textbf{ResNet18.} A popular vision model \cite{he2016deep}, ResNet18 is tested here using SatBird’s approach: we upsample the bioclimatic and pedological data to match RGB image resolution, concatenate the data as additional channels, and feed it through ResNet.
    \item \textbf{SatMAE.} SatMAE is a SOTA self-supervised model for satellite imagery \cite{cong2022satmae}, pretrained on Functional Map of the World and Sentinel-2 data \cite{christie2018functional}. Because SatMAE natively supports only satellite imagery, we test it in two configurations: (1) with frozen weights and satellite-only data, and (2) with environmental data integrated via Transformer, trained end-to-end. These are denoted SATMAE and SATMAE (img+env) respectively in Tables \ref{tab:summer} and \ref{tab:winter}.
    \item \textbf{SATLAS.} SATLAS, based on the Swin Transformer architecture, is another self-supervised satellite model \cite{bastani2023satlaspretrain} pretrained on the SATLAS dataset. Following SatBird’s approach, SATLAS is trained only with satellite imagery, as it lacks native support for environmental data. We test SATLAS in two modes: frozen weights and end-to-end fine-tuning, denoted SATLAS and SATLAS E2E respectively in Tables \ref{tab:summer} and \ref{tab:winter}.
\end{itemize}
Tables \ref{tab:summer} and \ref{tab:winter} summarize the results between the baseline and MiTREE model for both the SatBird Summer and SatBird Winter datasets. The MiTREE model can outperform the baseline models in all metrics in both summer and winter splits. Notably, in the summer split, \modelname achieves a top-10 score of 47.380\%, top-30 score of 66.609\%, and top-k score of 67.821\%, meaning it obtained the best performance in non-zero inflated metrics. Similarly, in the winter split, \modelname achieves the best top-10, top-30 and top-k at 51.773\%, 70.005\%, and 71.401\% respectively

\begin{figure*}[ht]
  \centering
  \vspace{5mm}
  \includegraphics[width=\linewidth]{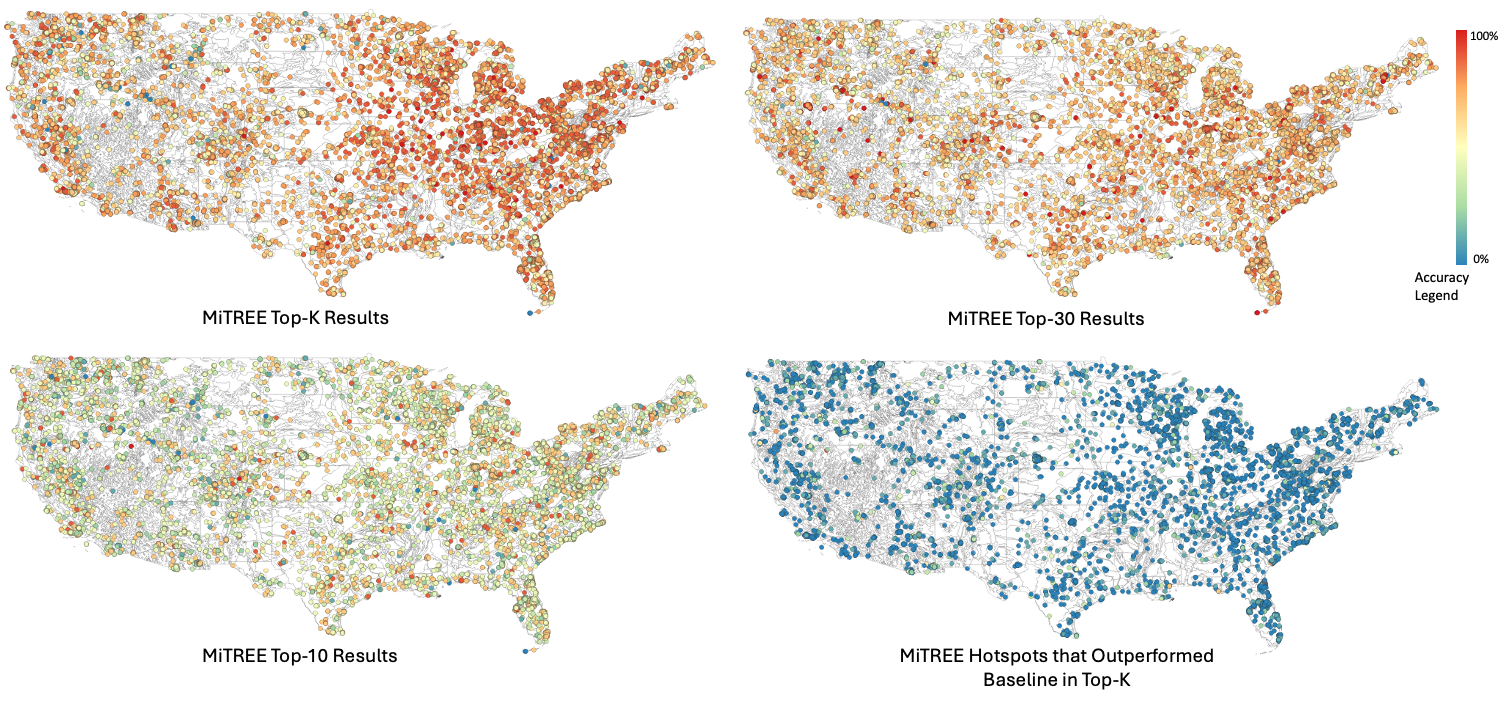}
  \caption{Visualizing the results for each test hotspots in the SatBird USA Summer split for \modelname. The points represent the results at the test hotspots, and the underlying map is the ecoregion polygons in the conterminous United States. For the outperformance map in the lower right, the color of the hotspots represents the difference in percent accuracy between \modelname and the ResNet baseline from (0, 100].}
  \label{fig:test_hotspots_mitree}
\end{figure*}
\begin{figure*}[ht]
  \centering
  \vspace{5mm}
  \includegraphics[width=\linewidth]{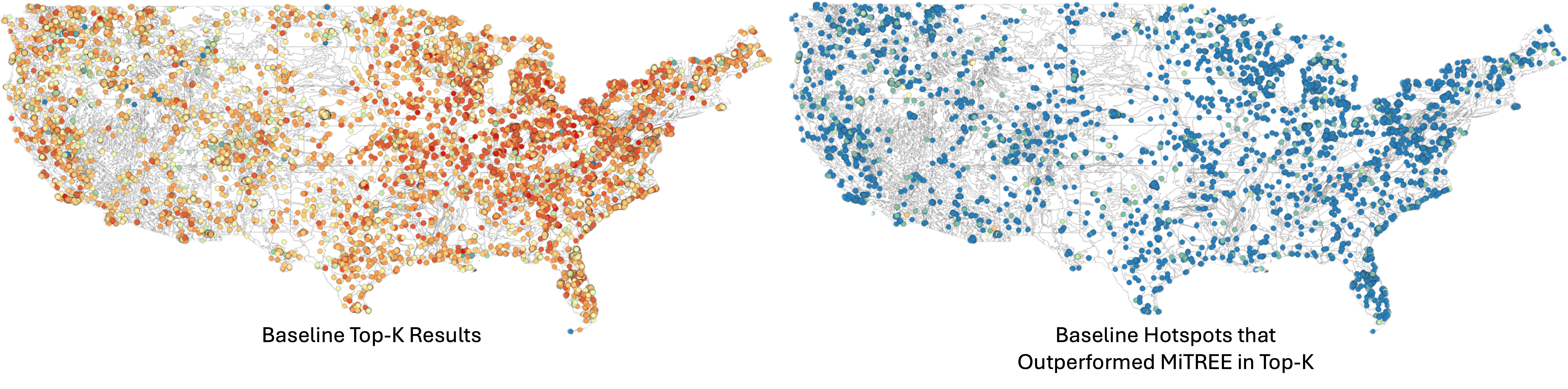}
  \caption{Visualizing the results for each test hotspots in the SatBird USA Summer split for the best performing baseline (ResNet). For the outperformance map on the right, the color of the hotspots represents the difference in percent accuracy between the ResNet baseline and the \modelname from (0, 100]. The colors of the hotspots denote the accuracy, which follows the same scale presented in Figure \ref{fig:test_hotspots_mitree}.}
  \label{fig:baseline_maps}
\end{figure*}

\subsection{Location Based Result Analysis}
\modelname results for each test hotspot in the SatBird Summer dataset are visualized in Figure \ref{fig:test_hotspots_mitree}. We choose to showcase summer results because there are more test hotspots than in the SatBird Winter split, and thus, there are more geographic locations represented.
\begin{figure}[h]
  \centering
  \vspace{5mm}
  \includegraphics[width=\linewidth]{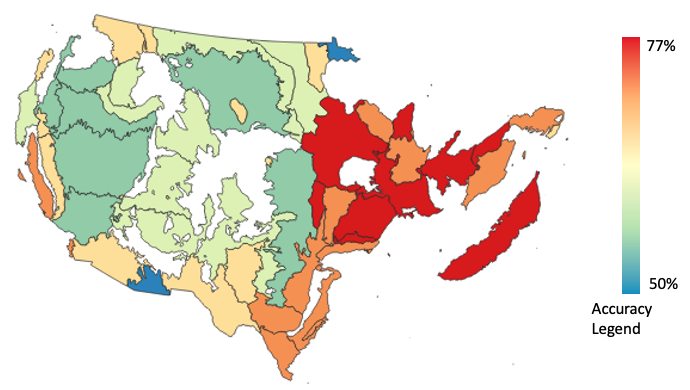}
  \caption{Top-k Accuracy by ecoregion for test hotspots in SatBird Summer over the ecoregion III map. Only ecoregions with test hotspots in them are shown. The accuracy legend is from 50 to 77 \% as this is the range of the accuracy per ecoregion.}
  \label{fig:ecoregionacc}
\end{figure}
\begin{figure}[h]
  \centering
  \vspace{5mm}
  \includegraphics[width=\linewidth]{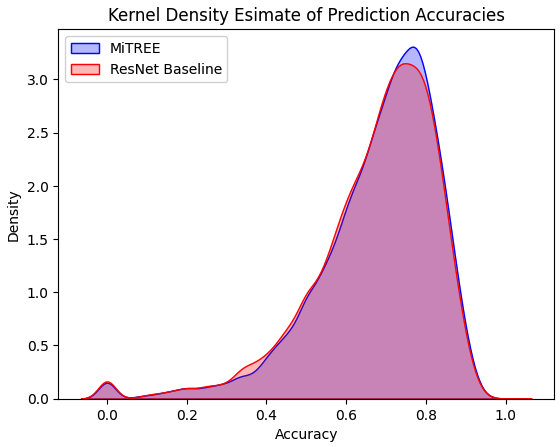}
  \caption{Visualizing the density of prediction accuracies for top-k accuracy between the baseline and MiTREE models.}
  \label{fig:kerneldensity}
\end{figure}

The prediction results for the hotspots are distributed relatively evenly across the United States. In the map for the top-k results, there is slightly higher accuracy in hotspots around the Midwest/Northeast region. This may be due to more sightings and higher encounter rates in that area. For instance, areas of relatively lower top-k accuracy are the northern Great Plains region in the middle of the United States. Those areas have fewer people and thus fewer observations, which makes the prediction of encounter rate much harder there. In the top-30 results, there is a similar cluster of high-accuracy hotspots in the Midwest/Northeast area and a reduction of accuracy in the northern Great Plains, likely reflecting the effect of having a lower birdwatching population in the area. The top-10 results are more difficult to decipher due to their having less high accuracy points, but it is possible to visually pick out a light orange higher accuracy cluster in the Northeast area and a light blue low accuracy cluster in the Great Plains region.

Compared to the best-performing baseline model (ResNet with concatenated img + env), \modelname obtains a better top-k result in 41\% (7,661 hotspots out of 18,529) of total hotspots in the summer dataset. The baseline obtains a better result in 32\% (6,008) of summer hotspots. The remaining 27\% (4900) of summer hotspots are ties. Overall, \modelname outperforms the baseline across a broader range of locations.

In Figure \ref{fig:test_hotspots_mitree} and Figure \ref{fig:baseline_maps}, the lower right-hand maps showcase hotspots where each model beat the other in top-k accuracy. As with the top-k result maps, there is no strong preference for a particular geographic area, though the \modelname outperformance map shows denser clusters simply because \modelname exceeds the baseline in more hotspots overall. Most differences in accuracy are small, as demonstrated by the blue points in the outperformance maps. Hotspots with large accuracy differences can be identified with a light red or green color and usually occur where there are very few species encounters (i.e., 1 or 2), so either one of the models can get a large accuracy difference if the other fails to predict even 1 species correctly. 

We analyze the types of hotspots that \modelname improves on by visualizing the density of the prediction top-k accuracies in Figure \ref{fig:kerneldensity}. The kernel density graph demonstrates that \modelname improves mid-range accuracies (30-40\%) to high accuracy predictions (70-90\%). This suggests that \modelname provides a robust representation of the environmental conditions at each hotspot. Less robust representations may only achieve moderate accuracy across a broader range of locations without fully capturing the underlying species distribution. A strong representation that effectively captures the ecological niche at a location should consistently predict most species present at that hotspot. 
\begin{figure*}[ht]
  \centering
  \vspace{5mm}
  \includegraphics[width=\linewidth]{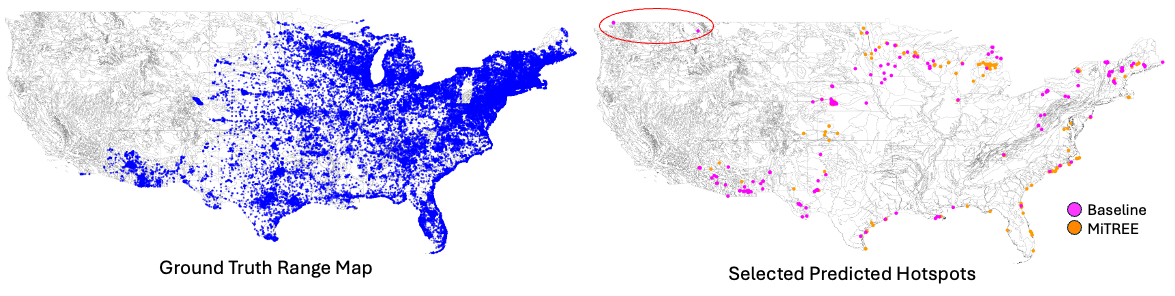}
  \caption{Visualizing the ground truth (left, blue) and predicted (right, orange and magenta) range for \emph{Cardinalis Cardinalis}, commonly known as the Northern Cardinal. The predicted hotspots on the right are where only one model predicted the hotspot. If the hotspot is magenta, that means it was only predicted by the baseline, otherwise if the hotspot is orange, it was only predicted by \modelname. The predicted range map was generated by plotting all hotspots with a predicted encounter rate over 10\%.}
  \label{fig:cardinal}
\end{figure*}

When computing the top-k accuracy by ecoregion, we find that most ecoregions have similar top-k accuracy (Figure \ref{fig:ecoregionacc}). Ecoregions with higher accuracy generally have more hotspots inside of them. All ecoregions in the ecoregion III map hotspots within them have a top k accuracy greater than 50\%. This indicates to us that \modelname performs consistently well across different ecoregions. This robustness suggests that the model is can make accurate predictions in various ecological contexts.

\subsection{Species Based Result Analysis}
Out of 670 species, \modelname obtains better average MSE on exactly 500, as shown in Table \ref{tab:species_compare}.
\begin{table}[ht]
\centering
\caption{Species-wise comparison between MiTREE and the baseline. The Mean MSE-D represents the average difference in MSE for species where the target model has outperformed the other model. For instance, the 0.000259 in the \modelname row indicates that, on average, if \modelname beats the baseline for some species, the MSE will be lower by 0.000259. The Max MSE-D represents the maximum MSE per species by which the target model has beaten the other model by.}
\label{tab:species_compare}
\begin{tabular}{lcccccc}
\toprule
Model Name & Mean MSE-D & Max MSE-D & \# Species \\
\midrule
Baseline & .000054 & .000842 & 170 \\
MiTREE & .000259 & .004026 & 500  \\
\bottomrule
\end{tabular}
\end{table}

The species-wise comparison in Table \ref{tab:species_compare} indicates that, in the cases where \modelname does outperform the baseline, it does so by a more significant margin than the baseline's improvement over \modelname. It is important to note that the Mean Squared Error (MSE) values tend to be small across all species due to zero inflation. 

Despite the overall small MSE values, the difference in mean MSE-D is still noteworthy. The higher mean MSE-D for \modelname means that when it outperforms the baseline, it does so by a higher margin, even in a context where improvements are difficult to achieve. The fact that \modelname consistently achieves larger improvements suggests that it is more effective than the baseline at capturing the underlying patterns in the data, leading to better predictive performance across a larger number of species. 

\subsubsection{Qualitative Analysis}
In order to better understand the differences between the performance of \modelname and the baseline, we visualized several ground truth species range maps alongside the predicted range maps. 

In Figure \ref{fig:cardinal}, we show the ground truth range map and a range map of select predictions for \emph{Cardinalis Cardinalis}, which is one of the species that \modelname had the highest outperformance compared to baseline. Most species with a large outperformance had very high occurrence rates in the training set. This may be because high occurrence rate species contribute more consistent patterns in the training set. 

One area to pay attention to in Figure \ref{fig:cardinal} is the region circled in red on the right-hand map. The two magenta dots represent two hotspots where the baseline model predicted a cardinal encounter rate of over 10\% but \modelname did not. Notably, these two dots are very far away from the rest of the predictions as well as well outside the ground truth range map. This could be a situation in which the ecoregion location encoder is helpful in differentiating the true range if the other inputs, like satellite imagery and environmental variables, are similar to other hotspots where the species habitates.

\section{Ablation Studies}

\subsection{Model Architecture}

We investigate the effect of the ResNet patch embedding layer on satellite imagery and the ecoregion location encoder. When swapping out the ResNet patch embedding layer, we substitute it with a patch embedding layer that is a single convolutional network, following the structure of a traditional ViT, and use Xavier initialization. For the ablation studies involving the ecoregion location encoder, we simply remove the ecoregion token from the model. We experiment with all combinations of these two modifications and demonstrate the results in Table \ref{tab:ablation}.

We find that the loss of the ResNet patch embedding layer has the greatest effect on accuracy. This is potentially because the pretrained (see Table \ref{tab:ablation_pretrain} for the effects of pretraining weights on model accuracy), deeper model can more effectively condense noisy satellite imagery. 

The ecoregion ablation study shows that including the ecoregion encoder marginally improved all metrics. Due to the small increase in performance, we also perform the ecoregion ablation study on the SatBird Winter dataset and find that the \modelname version with the ecoregion encoder also outperformed the non-ecoregion encoder version in every metric (Table \ref{tab:ablation_winter}).

\begin{table}[ht]
\centering
\caption{Ablation studies on the SatBird USA Summer dataset looking at the effect of the ResNet patch embedding layer and the ecoregion location encoder on the performance of the model. RN PE stands for ResNet Patch Embedding and Eco LE stands for Ecoregion Location Encoder.}
\label{tab:ablation}
\begin{tabular}{ccccccc}
\toprule
RN PE & Eco LE & MAE & MSE & Top-10 & Top-30 & Top-k \\
\midrule
\ding{55} & \ding{55} & 2.18 & .65 & 45.45\% & 64.84\% & 66.40\% \\
\ding{51} & \ding{55} & 2.08 & .63 & 46.99\% & 66.31\% & 67.55\% \\
\ding{55} & \ding{51} & 2.17 & .66 & 45.58\% & 65.34\% & 66.76\% \\
\ding{51} & \ding{51} & \textbf{2.07} & \textbf{.63} & \textbf{47.38\%} & \textbf{66.61\%} & \textbf{67.82\%}  \\
\bottomrule
\end{tabular}
\end{table}

\begin{table}[ht]
\centering
\caption{Ablation studies on the SatBird USA Winter dataset looking at the effect of the ecoregion location encoder on the performance of the model. Eco LE stands for ecoregion location encoder. Both models include the ResNet patch embedding.}
\label{tab:ablation_winter}
\begin{tabular}{ccccccc}
\toprule
Eco LE & MAE & MSE & Top-10 & Top-30 & Top-k \\
\midrule
\ding{55} & 1.82 & .44 & 51.73\% & 69.93\% & 71.36\% \\
\ding{51} & \textbf{1.62} & \textbf{.43} & \textbf{51.77\%} & \textbf{70.01\%} & \textbf{71.40\%} \\
\bottomrule
\end{tabular}
\end{table}

\subsection{Patch Embedding Initializations}
To investigate whether the improvement from the ResNet patch embedding could be outperformed by patch embeddings from SOTA ViT-based satellite image models, we conduct an ablation study looking at different types of initializations for the patch embedding of the satellite imagery input. We replace the patch embedding with a regular convolutional layer but loaded in weights from a pretrained SatMAE. To test the use of SATLAS as patch embedding initializations, we use the first layer of a SATLAS as the patch embedding. We had to use the first layer of SATLAS because SATLAS model uses a Swin Transformer as the backbone and has no convolution patch embedding layer. We present the results of the patch embedding initialization ablations in Table \ref{tab:ablation_pretrain}.

We find that loading the patch embedding layer with pretrained weights did not improve the results beyond MiTREE or even beyond the model with non-preloaded weights. Because the convolutional patch embedding layers are shallow, it is likely that most of the representation power in the original SatMAE and SATLAS models originated from their deeper Transformer layers.
\begin{table}[ht]
\centering
\caption{Ablation studies on the SatBird USA Summer dataset looking at the effect of loading different types of pretrained weights into the patch embedding layer for satellite imagery.}
\label{tab:ablation_pretrain}
\begin{tabular}{lcccccc}
\toprule
PE Weights & MAE & MSE & Top-10 & Top-30 & Top-k \\
\midrule
SatMAE PE & 2.25 & .68 & 43.40\% & 63.38\% & 65.06\% \\
Satlas PE & 2.20 & .66 & 44.32\% & 64.01\% & 65.72\% \\
Pretrain ResNet & \textbf{2.07} & \textbf{.63} & \textbf{47.38\%} & \textbf{66.61\%} & \textbf{67.82\%} \\
\bottomrule
\end{tabular}
\end{table}

\subsection{Ecoregions}
Domain experts and modelers have developed many variations of ecoregion maps, each with a distinct geographic scale and level of detail. To assess the impact of different ecoregion maps on model performance, we conducted experiments with four hierarchical levels of ecoregion maps, as shown in Table \ref{tab:ablation_ecoregions}.  The coarsest of these, Ecoregion Level I, represents large, broadly defined ecological areas. With each successive level—up to the most detailed Ecoregion Level IV—the regions become progressively smaller, capturing increasingly fine-grained ecological distinctions.

\begin{table}[ht]
\centering
\caption{Ablation studies on the SatBird USA Summer dataset looking at the effect of using different ecoregion maps with MiTREE. Ecoregion is abbreviated ER (i.e. ER I is equivalent to Ecoregion I).}
\label{tab:ablation_ecoregions}
\begin{tabular}{lcccccc}
\toprule
Ecoregion & MAE & MSE & Top-10 & Top-30 & Top-k \\
\midrule
ER I & 2.085 & .633 & 47.362\% & 66.608\% & \textbf{67.827\%} \\
ER II & 2.089 & .636 & 47.341\% & 66.541\% & 67.776\% \\
ER III & \textbf{2.070} & .630 & \textbf{47.380\%} & \textbf{66.609\%} & 67.821\% \\
ER IV & 2.086 & \textbf{.624} & 47.345\% & 66.544\% & 67.792\% \\
\bottomrule
\end{tabular}
\end{table}

The results of our ablation study show that all ecoregion maps yield comparable model performance across metrics, with Ecoregion Level III achieving a slight advantage by outperforming in three metrics: MAE, top-10, and top-30. This close performance across map levels may be explained by differing species' sensitivities to spatial scale: while some species benefit from coarse ecological labels that capture broad habitat types, others may have a smaller range and respond to finer-scale environmental distinctions. Consequently, each map level involves a trade-off that may favor certain species while limiting others.

Overall, our findings suggest that the inclusion of any ecoregion-based location and ecological indicator, even at a coarse level, improves the accuracy of species occurrence predictions. These results emphasize the value of incorporating ecological context to capture the spatial and environmental dependencies critical to species distribution modeling.

\section{Conclusion}
We introduce \modelname, a novel architecture that integrates a multi-input ViT with an ecoregion encoder, enabling the joint training of satellite imagery and environmental data without upsampling the input data. Our approach outperforms the best existing baselines on the SatBird Summer and Winter datasets, demonstrating the effectiveness of our modifications for the task of species distribution modeling. Additionally, our ablation studies confirm that the inclusion of our modifications- the ResNet patch embedding layer and the ecoregion encoder- significantly improves model performance across multiple metrics. 

Our intent with \modelname is to propose an effective model architecture for integrating environmental data and satellite imagery to represent geographic areas for ecological tasks. The \modelname framework is flexible and can be adapted for a variety of ecological downstream applications beyond bird species modeling. We believe this study accelerates connections between deep learning and species distribution modeling as well as ultimately advances the integration of deep learning into ecological tools.

For future work, we aim to investigate ways to incorporate a temporal element into our model architecture. We currently focus on spatial prediction only, but bird species distributions can be strongly affected by factors like climate change over time or seasonal variation. 

\section{Acknowledgements}
This research is part of AI-CLIMATE: “AI Institute for Climate-Land Interactions, Mitigation, Adaptation, Tradeoffs and Economy,” and is supported by USDA National Institute of Food and Agriculture (NIFA) and the National Science Foundation (NSF) National AI Research Institutes Competitive Award no. 2023-67021-39829. This material is also based upon works supported by the Defense Advanced Research Projects Agency (DARPA) under Agreement No. HR00112390132 and Contract No. 140D0423C0093. Any opinions, findings and conclusions or recommendations expressed in this material are those of the authors and do not necessarily reflect the views of the Defense Advanced Research Projects Agency (DARPA); or its Contracting Agent, the U.S. Department of the Interior, Interior Business Center, Acquisition Services Directorate, Division V. We would also like to acknowledge the support provided by the University of Minnesota College of Science and Engineering Fellowship.
\bibliographystyle{ACM-Reference-Format}
\bibliography{sample-base}

\end{document}